\documentclass{article} 
\usepackage{nips15submit_e,times}
\usepackage{hyperref}
\usepackage{url}
\usepackage{graphicx}
\usepackage{color}

\title{Scan, Attend and Read: End-to-End Handwritten Paragraph Recognition with MDLSTM Attention}

\author{
Th\'eodore Bluche  \qquad J\'er\^ome Louradour \qquad Ronaldo Messina \\
A2iA SAS\\
39 rue de la Bienfaisance\\
75008 Paris \\
\texttt{\{tb,rm\}@a2ia.com} \\
}

\nipsfinalcopy 

\begin{document}

\maketitle

\newcommand{\ie}{\textit{i.e.} }
\newcommand{\eg}{\textit{e.g.} }
\newcommand{\x}{$\mathbf{x}$}
\newcommand{\p}{\textbf{}}
\newcommand{\bi}[1]{\textit{\textbf{#1}}}
\newcommand{\fig}[1]{Figure~\ref{fig:#1}}
\newcommand{\sect}[1]{Section~\ref{sec:#1}}
\newcommand{\tab}[1]{Table~\ref{tab:#1}}
\newcommand{\eqn}[1]{Eqn.~\ref{eqn:#1}}

\begin{abstract}
We present an attention-based model for end-to-end handwriting recognition.
Our system does not require any segmentation of the input paragraph. 
The model is inspired by the differentiable attention models presented recently 
for speech recognition, image captioning or translation. The main difference is
the implementation of covert and overt attention with a multi-dimensional LSTM network. 
Our principal contribution towards handwriting recognition lies in the automatic 
transcription without a prior segmentation into lines, which was critical in previous approaches. 
To the best of our knowledge this is the first successful attempt 
of end-to-end multi-line handwriting recognition.
We carried out experiments on the well-known IAM Database. The results are encouraging
and bring hope to perform full paragraph transcription in the near future.
\end{abstract}

\section{Introduction}

In offline handwriting recognition, the input is a variable-sized two dimensional
image, and the output is a sequence of characters. The cursive 
nature of handwriting makes it hard to first segment characters to recognize 
them individually. Methods based on isolated characters were widely used in the nineties~\cite{bengio1995lerec,knerr1998hidden},
and progressively replaced by the sliding window approach, in which features are 
extracted from vertical frames of the line image~\cite{kaltenmeier1993sophisticated}. 
This method transforms the problem into a sequence to sequence transduction one, 
while potentially encoding the two-dimensional nature of the image by using 
convolutional neural networks~\cite{bluche2013feature} 
or by defining relevant features~\cite{Bianne_etal}.

The recent advances in deep learning and the new architectures allowed to build
systems that can handle both the 2D aspect of the input and the sequential aspect
of the prediction. In particular, Multi-Dimensional Long Short-Term Memory
Recurrent Neural Networks (MDLSTM-RNNs~\cite{Graves_Schmidhuber2008}),           
associated with the Connectionist Temporal Classification (CTC~\cite{Graves2006a}) objective function, 
yield low error rates and became the state-of-the-art model for handwriting recognition, 
winning most of the international evaluations in the field~\cite{maurdor,htrts,openhart}.  

Up to now, current systems require segmented text lines, which are rarely readily 
available in real-world applications. 
A complete processing pipeline must therefore
rely on automatic line segmentation algorithms in order to transcribe a document.
We propose a model for multi-line recognition,
built upon the recent ``attention-based'' methods, which have proven successful for 
machine translation~\cite{bahdanau2014neural}, 
image caption generation~\cite{cho2015describing,xu2015show}, 
or speech recognition~\cite{chan2015listen,chorowski2015attention}.
This proposal follows the longstanding and successful trend of making less and less segmentation 
hypotheses for handwriting recognition. 
Text recognition state-of-the-art moved from isolated character to isolated word recognition, then from isolated words to isolated lines recognition, and we now suggest to go further and recognize full pages without explicit segmentation.

Our domain of application bears similarities with the image captioning and
speech recognition tasks. We aim at selecting the relevant parts of an input signal to
sequentially generate text. Like in image captioning, the inputs are images. 
Similarly to the speech recognition task, we want to predict a monotonic and 
potentially long sequence of characters. 
In fact, we face here the challenges of both tasks.
We need an attention mechanism that should look for content at specific location and in 
a specific order. Moreover, in multi-line recognition, the reading order is 
encapsulated. For example, in Latin scripts, we have a primary order 
from left to right, and a secondary order from top to bottom. 
We deal here with a complex problem involving long two-dimensional sequences.

The system presented in this paper constitutes a whole new approach to handwriting 
recognition. Previous models make sequential predictions over the width of the 
image, with an horizontal step size fixed by the model. 
They have to resort to tricks to transform the 2D input image 
into a character sequence, such as sliding window and Hidden Markov Models, or
collapsing representations and CTC, making it impossible to handle multiple lines 
of text. Those approaches need the text to be already segmented into lines
to work properly. Moreover, the length of the predicted sequence, the reading 
order and the positions of predictions are directly embeded into the architecture.
Here, the sequence generation and extraction of information from the multi-dimensional 
input are decoupled. The system may adjust the number of predictions and arbitrarily
and iteratively select any part of the input.
The first results show that this kind of model could 
deprecate the need of line segmentation for training and recognition. 
Furthermore, since the model makes no assumption about the reading order, it could 
be applied without any change to languages with different reading order, such as 
Arabic (right-to-left, or even bidirectional when mixed with Latin scripts) or
some Asian languages (top-to-bottom).

\section{Handwriting Recognition with MDLSTM and CTC}
\label{sec:mdlstm}

\begin{figure}[ht]
\begin{center}
\includegraphics[width=\linewidth]{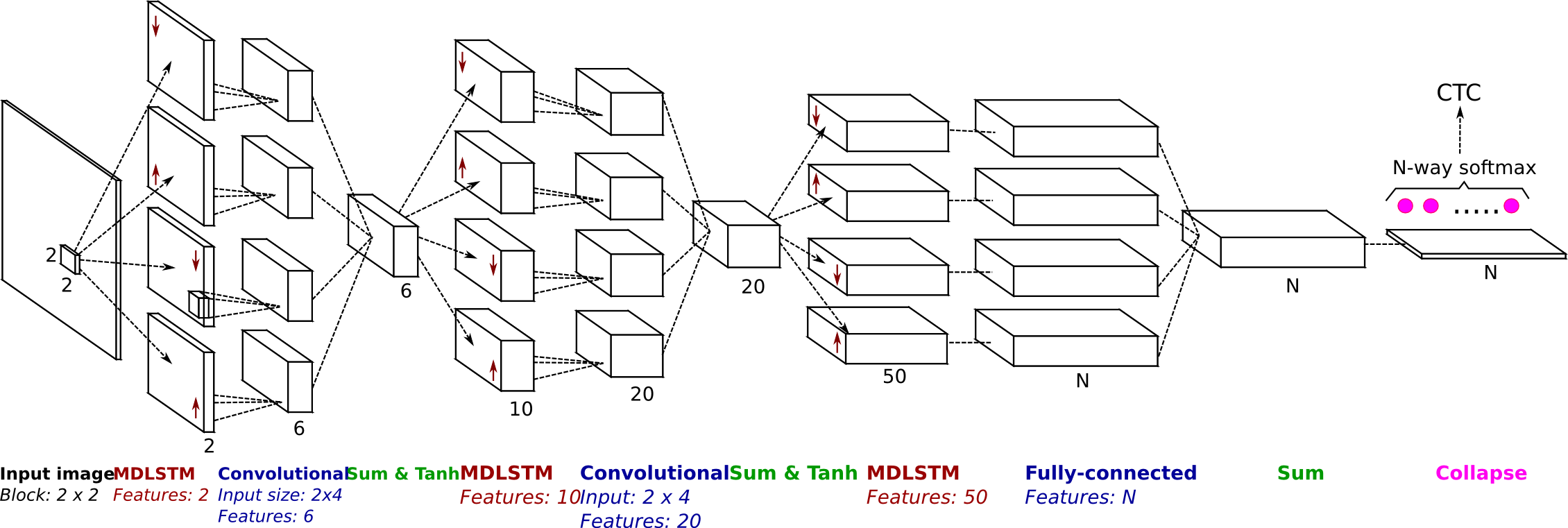}
\end{center}
\caption{MDLSTM-RNN for handwriting recognition, alternating LSTM layers in 
         four directions and subsampling convolutions. After the last linear layer,
         the feature maps are collapsed in the vertical dimension, and character
         predictions are obtained after a softmax normalization (figure from~\cite{pham2014dropout}).
         \label{fig:mdlstm}}
\end{figure}

Multi-Dimensional Long Short-Term Memory recurrent neural networks (MDLSTM-RNNs) 
were introduced in~\cite{Graves_Schmidhuber2008}      
for unconstrained handwriting recognition. They generalize the 
LSTM architecture to multi-dimensional inputs. An overview of the architecture 
is shown in \fig{mdlstm}. The image is presented to four MDLSTM layers, one layer for each
scanning direction. The LSTM cell inner state and output are computed from 
the states and output of previous positions in the horizontal and vertical directions:
\begin{equation}
 (h_{i,j}, q_{i,j}) = LSTM( x_{i,j}, h_{i,j \pm 1}, h_{i \pm 1, j}, q_{i,j \pm 1}, q_{i \pm 1, j})
\end{equation}
where $x_{i,j}$ is the input feature vector at position $(i,j)$, and $h$ and $q$ represent
the output and inner state of the cell, respectively. The $\pm 1$ choices in this
recurrence depend on which of the four scanning directions is considered.

Each LSTM layer is followed by a convolutional layer, with a step size greater than 
one, subsampling the feature maps. As in usual convolutional
architectures, the number of features computed by these layers increases as the size of 
the feature maps decreases. At the top of this network, there is one feature map 
for each label. A collapsing layer sums the features over the vertical axis,
yielding a sequence of prediction vectors, effectively delaying the 2D to 1D transformation 
just before the character predictions, normalized with a softmax activation.

In order to transform the sequence of $T$ predictions into a sequence of $N \leq T$
labels, an additionnal \textit{non-character} -- or blank -- label is introduced,
and a simple mapping is defined in order to obtain the final transcription.
The connectionist temporal classification objective (CTC~\cite{Graves2006a}), 
which considers all possible labellings of the sequence, is applied to train
the network to recognize a line of text.

The paradigm collapse/CTC already encodes the monotonicity of 
the prediction sequence, and allows to recognize characters from 2D images. 
In this paper, we propose to go beyond single line recognition, and to directly predict 
character sequences, potentially spanning several lines in the input 
image. To do this, we replace the collapse and CTC framework with an attention-based decoder.

\section{An Attention-Based Model for End-to-End Handwriting Recognition}

The proposed model comprises an encoder of the 2D image of text, producing 
feature maps, and a sequential decoder that predicts characters from these maps.
The decoder proceeds by combining the feature vectors of the encoded maps into 
a single vector, used to update an intermediate state and to predict the next 
character in the sequence. The weights of the linear combination of the feature
vectors at every timestep are predicted by an attention network. In this work
the attention is implemented with a MDLSTM network.

\begin{figure}[ht]
\begin{center}
\includegraphics[width=0.8\linewidth]{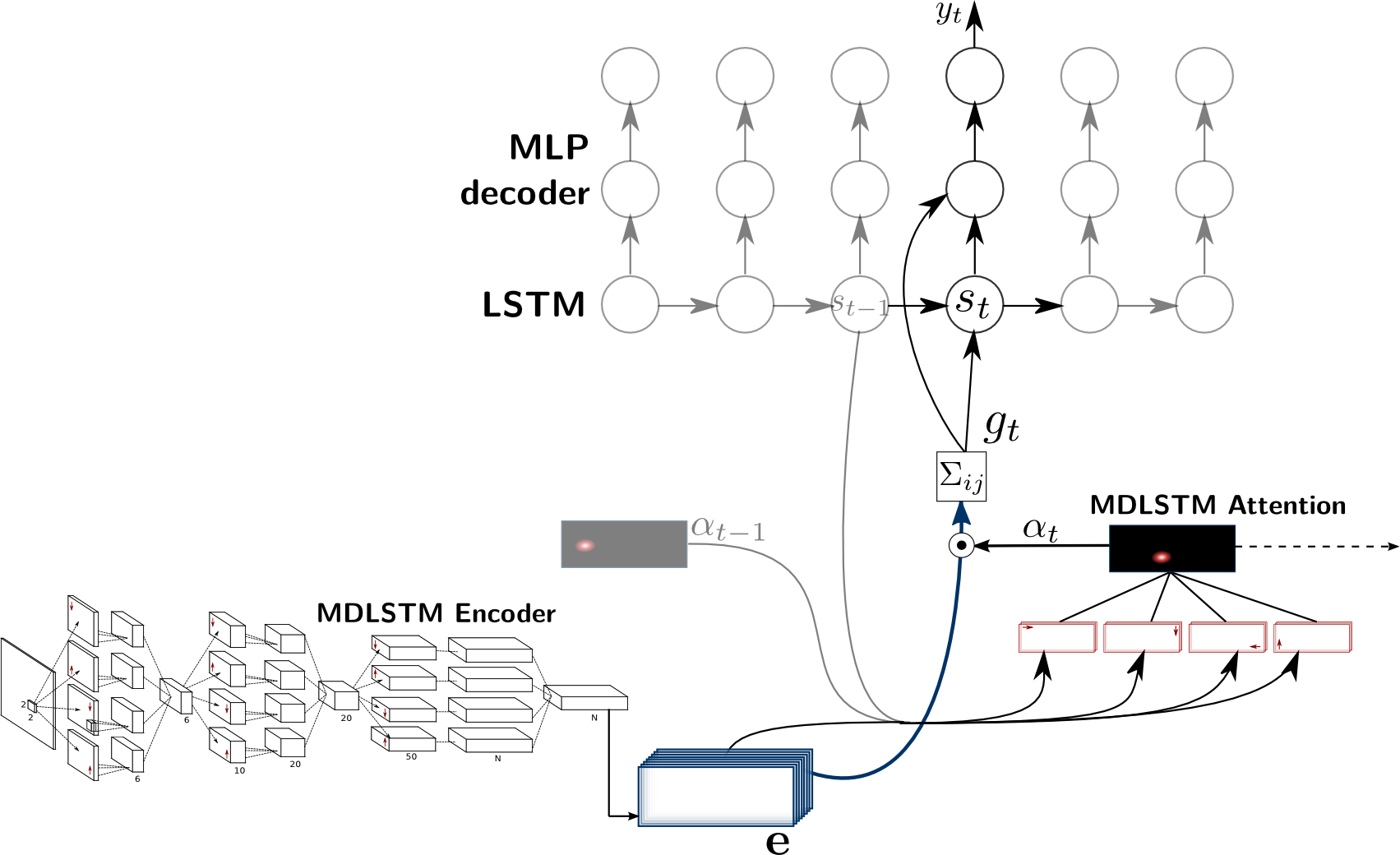}
\end{center}
\caption{Proposed architecture. The encoder network has the same architecture as
         the standard network of \fig{mdlstm}, except for the collapse and softmax
         layers. At each timestep, the feature maps, along with the previous attention
         map and state features are fed to an MDLSTM network which outputs new attention
         weights at each position. The weighted sum of the encoder features is computed
         and given to the state LSTM, and to the decoder. The decoder also considers the
         new state features and outputs character probabilities.\label{fig:archi}}
\end{figure}

The whole architecture, depicted in \fig{archi}, computes a fully differentiable
function, which parameters can be trained with backpropagation.
The optimized cost is the negative log-likelihood of the correct transcription:
\begin{equation}
 \mathcal{L}(\mathcal{I}, \mathbf{y}) = - \sum_{t} \log p(y_t | \mathcal{I})
\end{equation}
where $\mathcal{I}$ is the image, $\mathbf{y}=y_1,\cdots,y_T$ is the target character
sequence and $p( \cdotp | \mathcal{I})$ are the outputs of the network.


In the previous architecture (\fig{mdlstm}), we can see the MDLSTM network as a \textbf{feature extraction
module}, and the last collapsing and softmax layers as a way to predict sequences. 
Taking inspiration from \cite{cho2015describing,chorowski2015attention,xu2015show},
we keep the MDLSTM network as an encoder of the image $\mathcal{I}$ into high-level
features:
\begin{equation}
 e_{i,j} =  Encoder( \mathcal{I} )
\end{equation}
where $(i,j)$ are coordinates in the feature maps,
 and we apply an attention mechanism to read character from them.

\label{sec:attn_module}

The \textbf{attention mechanism} provides a summary of the encoded image at each timestep
in the form of a weighted sum of feature vectors. The attention network computes a 
score for the feature vectors at every position:
\begin{equation}
\label{eqn:attention}
 z_{(i,j), t} = Attention ( \mathbf{e}, \mathbf{\alpha}_{t-1}, s_{t-1} )
\end{equation}
We refer to $\mathbf{\alpha}_t = \{\alpha_{(i,j), t}\}_{(1 \leq i \leq W,~1 \leq j \leq H)}$
as the attention map at time $t$, which computation depends not only on the encoded
image, but also on the previous attention map, and on a state vector $s_{t-1}$.
The attention map is obtained by a softmax normalization:
\begin{equation}
 \alpha_{(i,j), t} = \frac{e^{z_{(i,j),t}}}{\sum_{i',j'} e^{z_{(i',j'),t}}}
\end{equation}

In the literature of attention-based models, we find two main kinds of mechanisms.
The first one is referred to as ``location-based'' attention. The attention network
in this case only predicts the position to attend from the previous attended position and
the current state (\eg in \cite{graves2013generating,graves2014neural}): 
\begin{equation}
 \alpha_{(i,j), t} = Attention ( \mathbf{\alpha}_{t-1}, s_{t-1} ) 
\end{equation}
The second kind of attention
is ``content-based''. The attention weights are predicted from the current state, and
the encoded features, \ie the network looks for relevant content (\eg in \cite{bahdanau2014neural,cho2015describing}):
\begin{equation}
 \alpha_{(i,j), t} = Attention ( \mathbf{e}, s_{t-1} ) 
\end{equation}
We combine these two complementary approaches to obtain the attention weights 
from both the content and the position, similarly to 
Chorowski et al.~\cite{chorowski2015attention}, who compute convolutional features 
on the previous attention weights in addition to the content-based features.

In this paper, we combine the previous attention map with the encoded features
through an MDLSTM layer, which can keep track of position and content (\eqn{attention}). 
With this architecture, the attention potentially depends on the context of the whole image.
Moreover, the LSTM gating system allows the network to use the content at one 
location to predict the attention weight for another location. In that sense, 
we can see this network as implementing a form of both overt and covert attention.

The state vector $s_t$ allows the model to keep track of what it has seen and done.
It is an ensemble of LSTM cells, whose inner states and outputs are 
updated at each timestep:
\begin{equation}
 s_t = LSTM( s_{t-1}, g_t )
\end{equation}
where $g_t$ represents the summary of the image at time $t$, resulting from the attention
given to the encoder features:
\begin{equation}
 g_t = \sum_{i,j} \alpha_{(i,j), t} e_{i,j}
\end{equation}
and is used both to update the state vector and to predict the next character.


The final component of this architecture is a \textbf{decoder}, which predicts the next character 
given the current image summary and state vector:
\begin{equation}
 y_t = Decoder( s_t, g_t )
\end{equation}
The end of sequence is predicted with a special token EOS.
In this paper, the decoder is a simple multi-layer perceptron with one hidden layer
($\tanh$ activation) and a softmax output layer.

\section{Related Work}

Our system is based on the idea of~\cite{bahdanau2014neural} 
to learn to align and transcribe for machine translation. It is achieved by coupling an encoder of the input signal and 
a decoder predicting language tokens with an attention mechanism, which selects 
from the encoded signal the relevant parts for the next prediction. 

It bears many similarity with the attention-based models for speech recognition
\cite{chan2015listen,chorowski2015attention}. Indeed, we want 
to predict text from a sensed version of natural language (audio in speech recognition,
image of handwritten text here). As for speech recognition, we need to deal with long sequences.
Our network also has LSTM recurrences, but we use MDLSTM units to handle images, 
instead of bi-directional LSTMs. This is a different way of handling images, 
compared with the attention-based systems for image captioning for example~\cite{cho2015describing,xu2015show}.
Besides the MDLSTM attention, the main difference in our architecture is that we do not
input the previous character to predict the next one, so it is also quite different
from the RNN transducers~\cite{Graves2012}.

Contrary to some attention models like DRAW~\cite{gregor2015draw} or spatial 
transformer networks~\cite{jaderberg2015spatial}, our model does not select and 
transform a part of the input by interpolation, but only weights the feature vectors
and combine them with a sum. We do not explicitely predict the coordinates of the
attention, as done in \cite{ba2014multiple}.

In similar models of attention, the weights are either computed from the content
at each position individually (e.g. in \cite{chan2015listen,xu2015show}), from
the location of the previous attention (e.g. in \cite{graves2013generating,graves2014neural})
or from a combination of both (e.g. in \cite{chorowski2015attention,graves2014neural}). 
In our model, the content of the whole image is explicitely taken into account
to predict the weight at every position, and the location is implicitely considered
through the MDLSTM recurrences. 

Finally, although attention models have been applied to the recognition of 
sequences of symbols (e.g. in~\cite{ba2014multiple,sonderby2015recurrent} for 
MNIST or SVHN digits, and~\cite{lee2016recursive,shi2016robust} for scene text
OCR on cropped words), we believe that we present the first attempt to recognize
multiple lines of cursive text without an explicit line segmentation.

\section{Experiments}

\subsection{Experimental Setup}

We carried out the experiments on the popular IAM database, described in details in~\cite{iam},
 consisting of images of handwritten English text documents. 
They correspond to English texts exctracted from the LOB corpus. 
657 writers produced between 1 and 59 handwritten documents. 
The training set comprises 747 documents (6,482 lines, 55,081 words), 
the validation set 116 documents (976 lines, 8,895 words) and the test set 
336 documents (2,915 lines, 25,920 words). The texts in this database typically
contain 450 characters in about nine lines. In 150 dpi images, the average character
has a width of 20px.

The baseline corresponds to the architecture presented in \fig{mdlstm},
 with 4, 20 and 100 units in MDLSTM layers, 12 and 32 units in convolutional layers, 
and dropout after every MDLSTM as presented in~\cite{pham2014dropout}. 
The last linear layer has 80 outputs, and is followed by a collapse layer and a
softmax normalization. In the attention-based model, the encoder has the same architecture
as the baseline model, without the collapse and softmax.
The attention network has 16 or 32 hidden LSTM units in each direction followed by
a linear layer with one output.
The state LSTM layer has 128 or 256 units, and the decoder is an MLP with 128 or 256
tanh neurons.
The networks are trained with RMSProp~\cite{tielmann} with a base learning rate of
$0.001$ and mini-batches of 8 examples. We measure the Character Error Rate (CER\%),
i.e. the edit distance normalized by the number of characters in the ground-truth.

\subsection{The Usual Word and Line Recognition Tasks}
We first trained the model to recognize words and lines.
The inputs are images of several consecutive words from the IAM database.
The encoder network has the standard architecture presented in \sect{mdlstm},
with dropout after each LSTM layer~\cite{pham2014dropout} and
was pre-trained on IAM database with CTC. 
The results are presented in \tab{tolines}. We see that the models tend to be better
on longer inputs, and the results for complete lines are not far from the baseline
performance.

\begin{table}[htb]
\begin{center}
\begin{minipage}{0.48\textwidth}
\caption{Multi-word recognition results (CER\%).\label{tab:tolines}}
\begin{center}
\begin{tabular}{|l|r|r|}\hline
\textbf{Model} & \textbf{Inputs} & \textbf{CER (\%)}  \\\hline
MDLSTM + CTC & Full Lines   & 6.6 \\\hline
Attention-based       & 1 word  & 12.6 \\
                      & 2 words & 9.4 \\
                      & 3 words & 8.2 \\
                      & 4 words & 7.8 \\
                      & Full Lines   & 7.0 \\\hline
 \end{tabular}
 \end{center}
\end{minipage}
\hfill
\begin{minipage}{0.48\textwidth}
\caption{Multi-line recognition results (CER\%).\label{tab:twolines}}
\begin{center}
\begin{tabular}{|r|r|}\hline
\textbf{Two lines of...} & \textbf{CER (\%)}  \\\hline
 1 words & 11.8 \\
 2 words & 11.1 \\
 3 words & 10.9 \\
 Full Lines & 9.4 \\\hline
 \end{tabular}
 \end{center}
\end{minipage}
 \end{center}
\end{table} 

\begin{figure}[ht]
\begin{center}
\includegraphics[width=\linewidth]{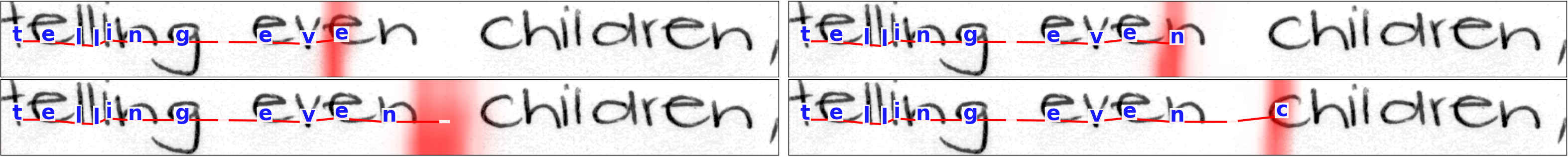}
\end{center}
\caption{Visualization of the attention weights at each timestep for multiple words. The attention 
map is interpolated to the size of the input image. The outputs of the network at 
each timestep are displayed in blue.\label{fig:attention4w}}
\end{figure}

In \fig{attention4w}, we display the attention map and character predictions 
as recognition proceeds. We see that attention effectively shifts from one 
character to the next, in the proper reading order.

\subsection{Learning Line Breaks}

Next, we evaluate the ability of this model to read multiple lines, \ie to 
read all characters of one line before finding the next line. This is 
challenging because it has to consider two levels of reading orders, which is crucial
to achieve whole paragraph recognition without prior line segmentation.

We started with a synthetic database derived from IAM, where the images of words
or sequences of words are stacked to represent two short lines.
The results (character error rate -- CER) are presented in \tab{twolines}.
Again, the system is better with longer inputs. The baseline from the previous 
section does not apply here anymore, and the error rate with two lines is worse than 
with a single line, but still in a reasonable range.

\begin{figure}[ht]
\begin{center}
\includegraphics[width=\linewidth]{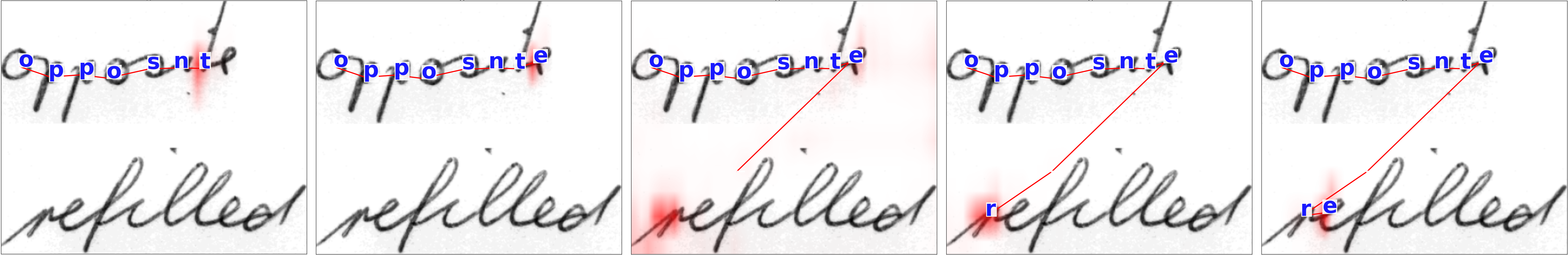}
\end{center}
\caption{Visualization of the attention weights at each timestep for multiple lines. The attention 
map is interpolated to the size of the input image.\label{fig:attention2wstack}}
\end{figure}

We show in \fig{attention2wstack} the outputs of the decoder and of the attention 
network on an example of two lines of one word. We observe that the system learnt to
look for the second line when the first line is read, with an attention split between
the end of the first line and the beginning of the second line.

\subsection{Towards Paragraph Recognition}

Training this system on paragraphs raises several challenges. The model still 
has to learn to both align and recognize, but the alignment problem is much more 
complex. A typical paragraph from IAM contains 450 characters on 9 text lines.
Moreover, the full backpropagation though time must cover those 450 timesteps,
on images that are significantly bigger than the line images, which is prohibitive
in terms of memory usage.

To tackle these challenges, we modified the training procedure in several ways.
First, we truncated the backpropagation through time of the decoder to 30 timesteps in 
order to adress the memory issue. Note that although 30 timesteps was chosen so that
intermediate activations fit in memory even for full paragraphs, it roughly corresponds 
to half a line, or 4-5 words, and we suppose that it is sufficient to learn the 
relevant dependencies. 
Then, instead of using only full paragraphs 
(there are only 747 in the training set), we added the single lines and all 
concatenations of successive lines. To some extent, this may be seen as data augmentation 
by considering different crops of paragraphs. 

Finally, we applied several levels of curriculum learning~\cite{Bengio2009}. One of these is the 
strategy proposed by~\cite{louradour}, which samples training examples according to their
target length. It prefers short sequences at the beginning of training (e.g. single lines)
and progressively adds longer sequences (paragraphs). The second curriculum is 
similar to that of~\cite{ba2014multiple}: we train only to recognize the first few
characters at the beginning. The targets are the first $N \times epoch$ characters, 
with $N=50$, i.e. first 50 during the first epoch, then first 100, and so on.
Note that 50 characters roughly correspond to the length of one line. 
This strategy amounts to train to recognize the first line during the first epoch,
then the first two lines, and so on.

The baseline here is the MDLSTM network trained with CTC for single lines,
applied to the result of automatic line segmentation. We present in \tab{autoseg}
the character error rates obtained with different input resolutions and segmentation
algorithms. Note that the line segmentation on IAM is quite easy as the lines 
tend to be clearly separated.

\begin{table}[htb]
\caption{Character Error Rates (\%) of CTC-trained RNNs on ground-truth lines and 
 automatic segmentation of paragraphs with different resolutions.\label{tab:autoseg}}
\begin{center}
\begin{tabular}{|r|c|c|c|c|c|}\hline
 \textbf{Resolution} &  \multicolumn{4}{c|}{\textbf{Line segmentation}} & \multicolumn{1}{r|}{\textbf{Attention-based}} \\\cline{2-5}
 \textbf{(DPI)} & \textbf{GroundTruth} & \textbf{Projection} & \textbf{Shredding} & \textbf{Energy} & \textit{(this work)} \\\hline
   90  & 18.8 & 24.7 & 19.8 & 20.8 & - \\ 
  150  & 10.3 & 17.2 & 11.1 & 11.8 & 16.2 \\ 
  300  &  6.6 & 13.8 &  7.5 &  7.9 & - \\\hline
 \end{tabular}
 \end{center}
\end{table}

\begin{figure}[ht]
\begin{center}
\includegraphics[width=0.4\linewidth]{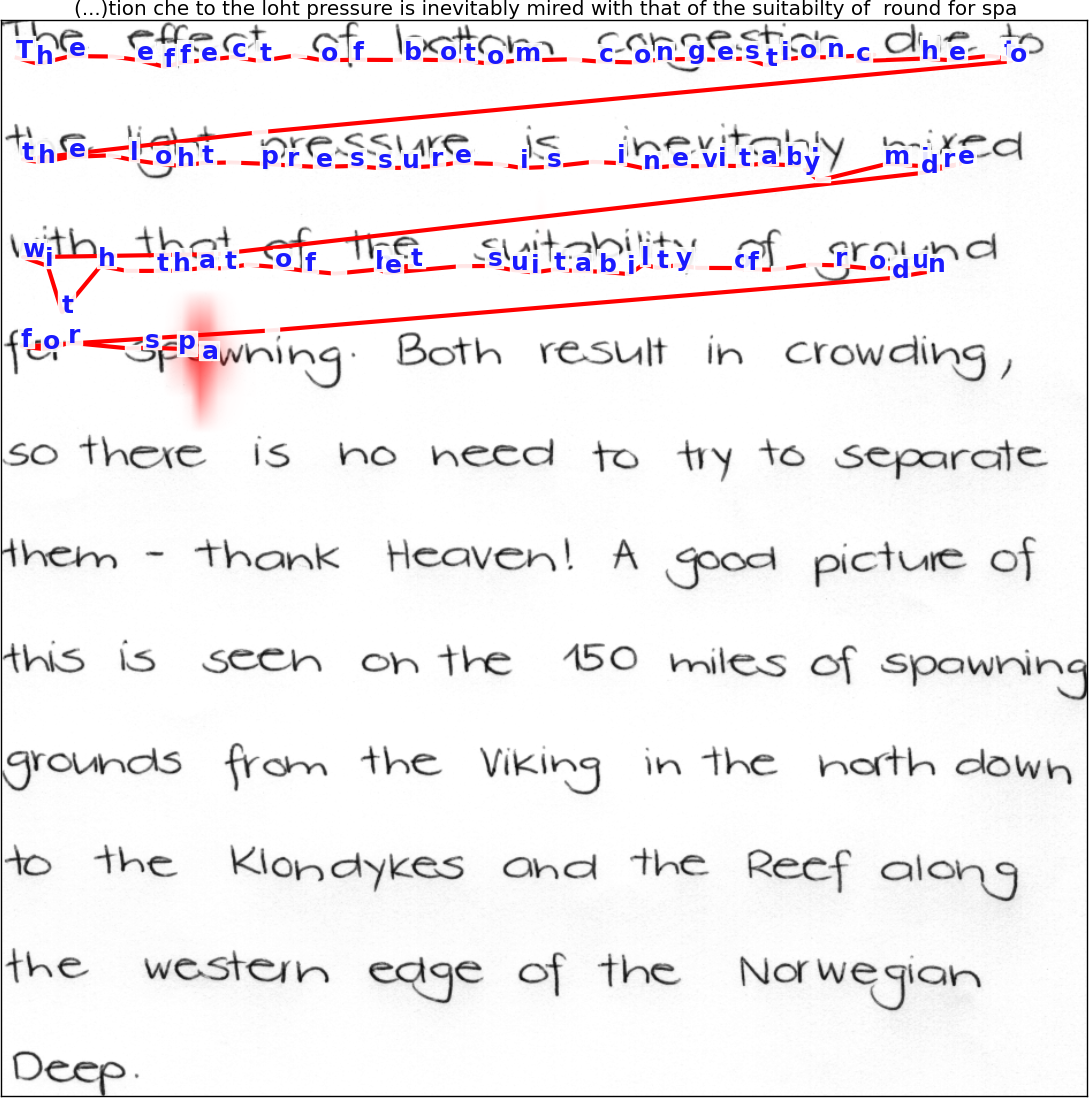}
\includegraphics[width=0.4\linewidth]{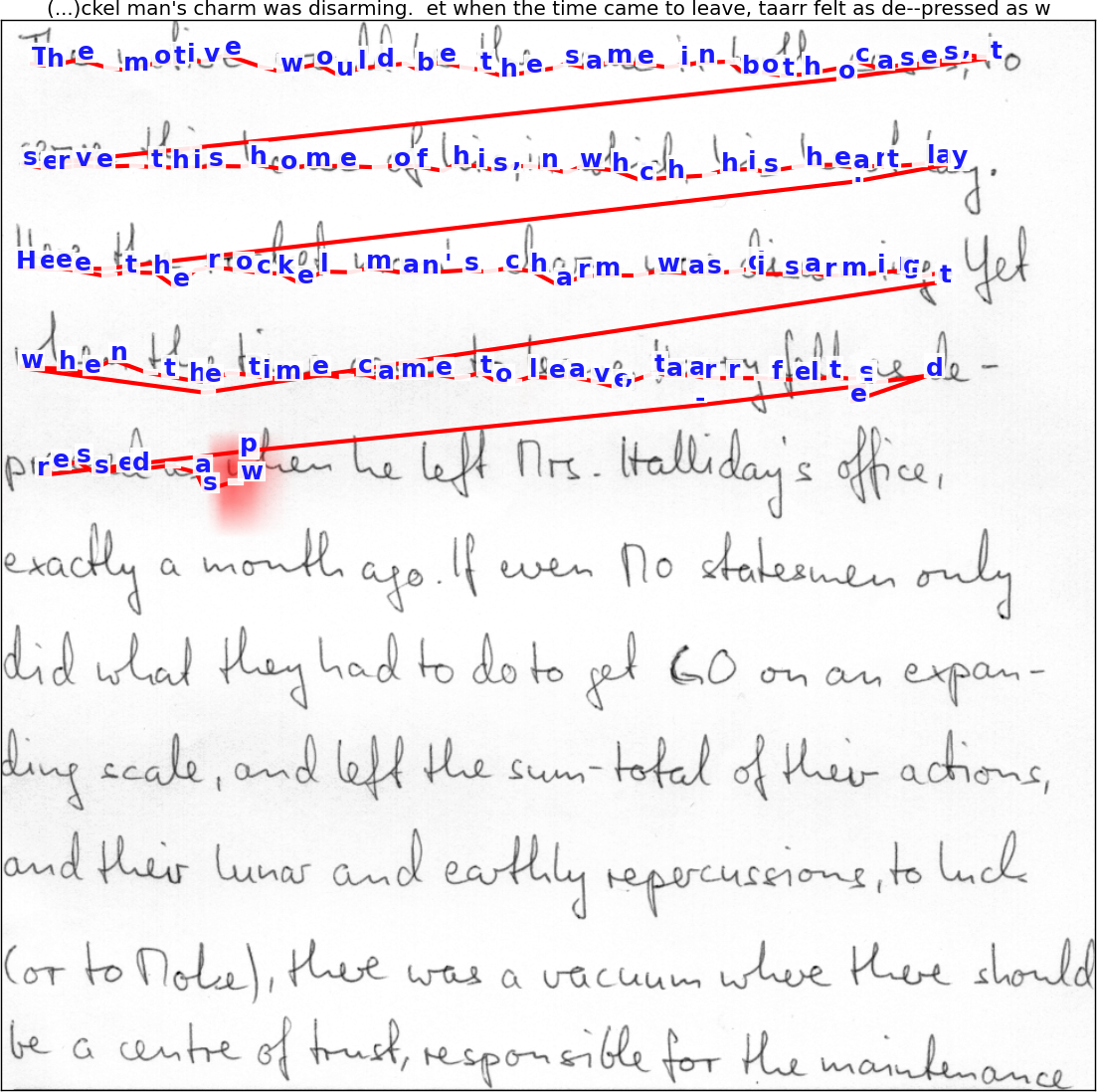} 
\caption{Transcribing full paragraphs of text. Character predictions are located 
at the center of mass of the attention weights.
An online demo is available at \url{https://youtu.be/qIs1SUH_3Lw}.
\label{fig:para}}
 \end{center}
\end{figure}

\begin{figure}[ht]
\begin{center}
\includegraphics[width=0.9\linewidth]{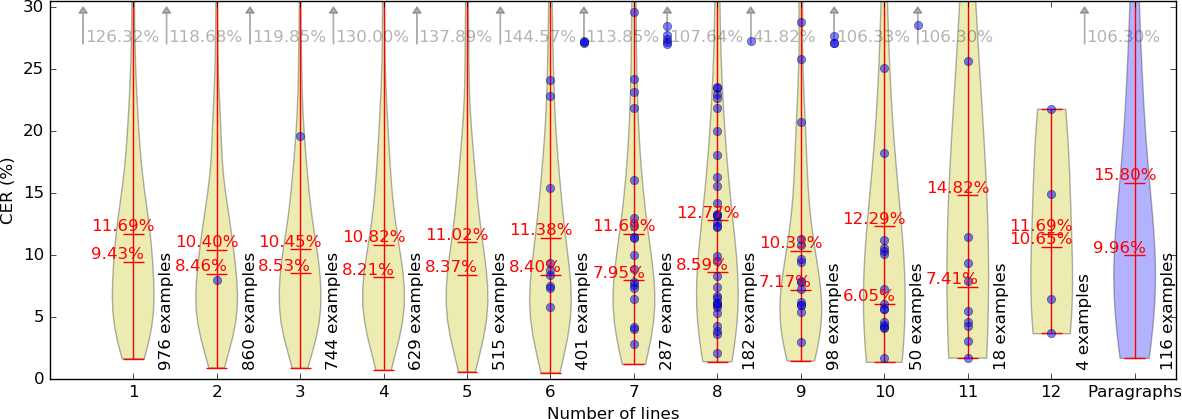}
\caption{Character Error Rates (\%) of the proposed model trained on multiple lines,
evaluated with inputs containing 
different number of lines (150 dpi, after twelve epochs).
The medians and means accross all examples are displayed in red.
The blue dots are complete paragraphs.\label{fig:paraseg}}
 \end{center}
\end{figure}

We trained the attention-based model on 150 dpi images and the results after only
twelve epochs are promising. In \fig{para}, we show some examples of paragraphs 
being transcribed by the network. We report the character error rates
on inputs corresponding to all possible sub-paragraphs of one to twelve lines 
from the development set in \fig{paraseg}. The \textit{Paragraphs} column corresponds 
to the set of actual complete paragraphs, individually depicted as blue dots in 
the other columns. Note that for a few samples, the 
attention jumped back to a previous line at some point, causing the system to 
transcribe again a whole part of the image. In those cases, the insertion rate
was very high and the final CER sometimes above 100\%. 

\section{Discussion}

The results we present in this paper are promising and show that recognizing 
full paragraphs of text without an explicit segmentation into lines is feasible.
Not only can we hope to perform full paragraph recognition in the near future,
but we may also envision the recognition of complex documents. The attention 
mechanism would then be a way of performing document layout analysis and 
text recognition within a single end-to-end system. 

We also carried out preliminary experiments on Arabic text lines and SVHN without 
any cropping, rescaling, or preprocessing. The results are interesting. For Arabic,
the model effectively reads from right to left, and manages to handle bidirectional
reading order in mixed Arabic/Latin inputs in several images. For SVHN, the model
finds digits in the scene images.


In this version of the model, the prediction is not explicitely conditioned on the 
previous character, as for example in~\cite{chorowski2015attention},
and the integration of a language model is more complicated than with classical models
trained with CTC. This should be addressed in future work. 
Finally, the presented system is very slow due to the computation of
 attention for each character in turn. The time and memory consumption is prohibitive 
for most industrial applications, but learning how to read whole paragraphs
might open new directions of research in the field.

\section{Conclusion}

In this paper, we have presented a method to transcribe complete paragraphs of
text without an explicit line segmentation. The system is based on MDLSTM-RNNs,
widely applied to transcribe isolated text lines, and is inspired from the 
recent attention-based models. The proposed model is able to recognize multiple
lines of text, and to learn encapsulated reading orders. It is not limited to 
handwritten Latin scripts, and could be applied without change to other languages
(such as Chinese or Arabic), write type (e.g. printed text), or more generally
image-to-sequence problems.

Unlike similar models, the decoder output is not conditioned on the previous token. 
Future work will include this architectural modification, which would enable 
a richer decoding with a beam search. 
On the other hand, we proposed an MDLSTM attention network, which computes attention
weights taking into account the context of the whole image, and merging location 
and content information.

The results are encouraging, and prove that explicit segmentation is not necessary,
which we believe is an important contribution towards end-to-end handwriting 
recognition.

%

\bibliographystyle{plain}
\small
\bibliography{sar}

\begin{thebibliography}{10}

\bibitem{ba2014multiple}
Jimmy Ba, Volodymyr Mnih, and Koray Kavukcuoglu.
\newblock Multiple object recognition with visual attention.
\newblock {\em arXiv preprint arXiv:1412.7755}, 2014.

\bibitem{bahdanau2014neural}
Dzmitry Bahdanau, Kyunghyun Cho, and Yoshua Bengio.
\newblock Neural machine translation by jointly learning to align and
  translate.
\newblock {\em arXiv preprint arXiv:1409.0473}, 2014.

\bibitem{bengio1995lerec}
Yoshua Bengio, Yann LeCun, Craig Nohl, and Chris Burges.
\newblock {Lerec: A NN/HMM hybrid for on-line handwriting recognition}.
\newblock {\em Neural Computation}, 7(6):1289--1303, 1995.

\bibitem{Bengio2009}
Yoshua Bengio, J\'er\^ome Louradour, Ronan Collobert, and Jason Weston.
\newblock Curriculum learning.
\newblock In {\em ICML}, page~6, 2009.

\bibitem{Bianne_etal}
A.-L. Bianne, F.~Menasri, R.~Al-Hajj, C.~Mokbel, C.~Kermorvant, and
  L.~Likforman-Sulem.
\newblock {Dynamic and Contextual Information in HMM modeling for Handwriting
  Recognition}.
\newblock {\em IEEE Trans. on Pattern Analysis and Machine Intelligence},
  33(10):2066 -- 2080, 2011.

\bibitem{bluche2013feature}
Th{\'e}odore Bluche, Hermann Ney, and Christopher Kermorvant.
\newblock {Feature Extraction with Convolutional Neural Networks for
  Handwritten Word Recognition}.
\newblock In {\em 12th International Conference on Document Analysis and
  Recognition (ICDAR)}, pages 285--289. IEEE, 2013.

\bibitem{maurdor}
Sylvie Brunessaux, Patrick Giroux, Bruno Grilh{\`e}res, Mathieu Manta, Maylis
  Bodin, Khalid Choukri, Olivier Galibert, and Juliette Kahn.
\newblock {The Maurdor Project: Improving Automatic Processing of Digital
  Documents}.
\newblock In {\em Document Analysis Systems (DAS), 2014 11th IAPR International
  Workshop on}, pages 349--354. IEEE, 2014.

\bibitem{chan2015listen}
William Chan, Navdeep Jaitly, Quoc~V Le, and Oriol Vinyals.
\newblock Listen, attend and spell.
\newblock {\em arXiv preprint arXiv:1508.01211}, 2015.

\bibitem{cho2015describing}
Kyunghyun Cho, Aaron Courville, and Yoshua Bengio.
\newblock Describing multimedia content using attention-based encoder-decoder
  networks.
\newblock {\em Multimedia, IEEE Transactions on}, 17(11):1875--1886, 2015.

\bibitem{chorowski2015attention}
Jan~K Chorowski, Dzmitry Bahdanau, Dmitriy Serdyuk, Kyunghyun Cho, and Yoshua
  Bengio.
\newblock Attention-based models for speech recognition.
\newblock In {\em Advances in Neural Information Processing Systems}, pages
  577--585, 2015.

\bibitem{Graves2006a}
A~Graves, S~Fern\'{a}ndez, F~Gomez, and J~Schmidhuber.
\newblock {Connectionist temporal classification: labelling unsegmented
  sequence data with recurrent neural networks}.
\newblock In {\em International Conference on Machine learning}, pages
  369--376, 2006.

\bibitem{Graves_Schmidhuber2008}
A.~Graves and J.~Schmidhuber.
\newblock {Offline Handwriting Recognition with Multidimensional Recurrent
  Neural Networks}.
\newblock In {\em Advances in Neural Information Processing Systems}, pages
  545--552, 2008.

\bibitem{Graves2012}
Alex Graves.
\newblock {Sequence Transduction with Recurrent Neural Networks}.
\newblock In {\em ICML}, 2012.

\bibitem{graves2013generating}
Alex Graves.
\newblock Generating sequences with recurrent neural networks.
\newblock {\em arXiv preprint arXiv:1308.0850}, 2013.

\bibitem{graves2014neural}
Alex Graves, Greg Wayne, and Ivo Danihelka.
\newblock Neural turing machines.
\newblock {\em arXiv preprint arXiv:1410.5401}, 2014.

\bibitem{gregor2015draw}
Karol Gregor, Ivo Danihelka, Alex Graves, and Daan Wierstra.
\newblock {DRAW: A recurrent neural network for image generation}.
\newblock {\em arXiv preprint arXiv:1502.04623}, 2015.

\bibitem{jaderberg2015spatial}
Max Jaderberg, Karen Simonyan, Andrew Zisserman, et~al.
\newblock Spatial transformer networks.
\newblock In {\em Advances in Neural Information Processing Systems}, pages
  2008--2016, 2015.

\bibitem{kaltenmeier1993sophisticated}
Alfred Kaltenmeier, Torsten Caesar, Joachim~M Gloger, and Eberhard Mandler.
\newblock {Sophisticated topology of hidden Markov models for cursive script
  recognition}.
\newblock In {\em Document Analysis and Recognition, 1993., Proceedings of the
  Second International Conference on}, pages 139--142. IEEE, 1993.

\bibitem{knerr1998hidden}
Stefan Knerr, Emmanuel Augustin, Olivier Baret, and David Price.
\newblock {Hidden Markov model based word recognition and its application to
  legal amount reading on French checks}.
\newblock {\em Computer Vision and Image Understanding}, 70(3):404--419, 1998.

\bibitem{lee2016recursive}
Chen-Yu Lee and Simon Osindero.
\newblock Recursive recurrent nets with attention modeling for ocr in the wild.
\newblock {\em arXiv preprint arXiv:1603.03101}, 2016.

\bibitem{louradour}
J\'er\^ome Louradour and Christopher Kermorvant.
\newblock Curriculum learning for handwritten text line recognition.
\newblock In {\em International Workshop on Document Analysis Systems (DAS)},
  2014.

\bibitem{iam}
U-V Marti and Horst Bunke.
\newblock {The IAM-database: an English sentence database for offline
  handwriting recognition}.
\newblock {\em International Journal on Document Analysis and Recognition},
  5(1):39--46, 2002.

\bibitem{pham2014dropout}
Vu~Pham, Th{\'e}odore Bluche, Christopher Kermorvant, and J{\'e}r{\^o}me
  Louradour.
\newblock {Dropout improves recurrent neural networks for handwriting
  recognition}.
\newblock In {\em 14th International Conference on Frontiers in Handwriting
  Recognition (ICFHR2014)}, pages 285--290, 2014.

\bibitem{htrts}
Joan~Andreu S\'anchez, Ver\'onica Romero, Alejandro Toselli, and Enrique Vidal.
\newblock {ICFHR 2014 HTRtS: Handwritten Text Recognition on tranScriptorium
  Datasets}.
\newblock In {\em International Conference on Frontiers in Handwriting
  Recognition (ICFHR)}, 2014.

\bibitem{shi2016robust}
Baoguang Shi, Xinggang Wang, Pengyuan Lv, Cong Yao, and Xiang Bai.
\newblock Robust scene text recognition with automatic rectification.
\newblock {\em arXiv preprint arXiv:1603.03915}, 2016.

\bibitem{sonderby2015recurrent}
S{\o}ren~Kaae S{\o}nderby, Casper~Kaae S{\o}nderby, Lars Maal{\o}e, and Ole
  Winther.
\newblock Recurrent spatial transformer networks.
\newblock {\em arXiv preprint arXiv:1509.05329}, 2015.

\bibitem{tielmann}
Tijmen Tieleman and Geoffrey Hinton.
\newblock Lecture 6.5-rmsprop: Divide the gradient by a running average of its
  recent magnitude.
\newblock {\em COURSERA: Neural Networks for Machine Learning}, 4, 2012.

\bibitem{openhart}
A.~Tong, M.~Przybocki, V.~Maergner, and H.~El~Abed.
\newblock {NIST 2013 Open Handwriting Recognition and Translation (OpenHaRT13)
  Evaluation}.
\newblock In {\em 11th IAPR Workshop on Document Analysis Systems (DAS2014)},
  2014.

\bibitem{xu2015show}
Kelvin Xu, Jimmy Ba, Ryan Kiros, Aaron Courville, Ruslan Salakhutdinov, Richard
  Zemel, and Yoshua Bengio.
\newblock Show, attend and tell: Neural image caption generation with visual
  attention.
\newblock {\em arXiv preprint arXiv:1502.03044}, 2015.

\end{thebibliography}
\end{document}